\documentclass[sigconf]{acmart}
\acmConference[ISSTA 2023]{ACM SIGSOFT International Symposium on Software Testing and Analysis}{17-21 July, 2023}{Seattle, USA}

\usepackage{url}
\usepackage{graphicx}
\usepackage{bm}
\usepackage{subfigure} 
\usepackage{array,color}
\usepackage{multirow}
\usepackage{ulem}
\usepackage{caption}
\usepackage[vlined,boxed,linesnumbered, ruled]{algorithm2e}
\usepackage{algorithmic}
\usepackage{stfloats}

\renewcommand\footnotetextcopyrightpermission[1]{} 

\AtBeginDocument{%
  }

\begin{document}
\title{HyperAttack: Multi-Gradient-Guided White-box Adversarial Structure Attack of Hypergraph Neural Networks}
\author{Chao Hu}
\affiliation{%
  \institution{Central South University}
  \city{Hunan}
  \country{China}
}

\author{Ruishi Yu}
\affiliation{
  \institution{Central South University}
  \city{Hunan}
  \country{China}
}

\author{Binqi Zeng}
\affiliation{
  \institution{Central South University}
  \city{Hunan}
  \country{China}
}

\author{Yu Zhan}
\affiliation{
  \institution{Central South University}
  \city{Hunan}
  \country{China}
}

\author{Ying Fu}
\affiliation{
  \institution{National University of Defence Technology}
  \city{Hunan}
  \country{China}
}
\author{Quan Zhang}
\affiliation{
  \institution{Tsinghua University}
  \city{Beijing}
  \country{China}
}

\author{Rongkai Liu}
\affiliation{
  \institution{Central South University}
  \city{Hunan}
  \country{China}
}

\author{Heyuan Shi}
\authornote{Corresponding author}
\affiliation{
  \institution{Central South University}
  \city{Hunan}
  \country{China}
}

\renewcommand{\shortauthors}{xxx et al.}

\begin{abstract}

Hypergraph neural networks (HGNN) have shown superior performance in various deep learning tasks, leveraging the high-order representation ability to formulate complex correlations among data by connecting two or more nodes through hyperedge modeling. Despite the well-studied adversarial attacks on Graph Neural Networks (GNN), there is few study on adversarial attacks against HGNN, which leads to a threat to the safety of HGNN applications. 
In this paper, we introduce HyperAttack, the first white-box adversarial attack framework against hypergraph neural networks. 
HyperAttack conducts a white-box structure attack by perturbing hyperedge link status towards the target node with the guidance of both gradients and integrated gradients.
We evaluate HyperAttack on the widely-used Cora and PubMed datasets and three hypergraph neural networks with typical hypergraph modeling techniques. 
Compared to state-of-the-art white-box structural attack methods for GNN, HyperAttack achieves a 10–20X improvement in time efficiency while also increasing attack success rates by 1.3$\%$-3.7$\%$. 
The results show that HyperAttack can achieve efficient adversarial attacks that balance effectiveness and time costs.



\end{abstract}







\maketitle


\section{Introduction}\label{sec1}


As a unique non-Euclidean data structure for machine learning, graph modeling and hypergraph modeling are widely used to formulate the relationships between data, and graph neural networks (GNN) and hypergraph neural networks (HGNNs) are designed to derive knowledge embedding and learn the low-dimensional representation of nodes and links in many deep learning tasks.
Various types of GNN have been proposed, e.g., GraphSAGE~\cite{hamilton2017inductive}, Graph Convolutional Networks (GCN)~\cite{KipfW16GCN} and Graph Attention Network (GAT)~\cite{VelickovicCCRLB18GAT}, for many graph-based analysis tasks, such as link prediction~\cite{perozzi2014deepwalk,wang2017signed}, node classification~\cite{tang2015pte,wang2016linked}, community detection~\cite{tian2014learning,allab2016semi}, and social network analysis~\cite{liu2016aligning}.
Though the network structure of GNNs is different, they all aggregate information based on graph modeling, where one edge can only have two nodes in a graph.
Hypergraph Neural Networks (HGNNs) have become a hot topic in recent years, and various HGNNs have been proposed, such as HyperGCN~\cite{Yadati19Hypergcn}, HGNN$^+$~\cite{Gao23HGNN+} and  Dynamic Hypergraph Neural Networks (DHGNN)~\cite{Jiang19DHNN}.
HGNNs take advantage of the stronger relational representation ability of hypergraph modeling, i.e., one hyperedge can connect two or more number of nodes, which is more practical to formulate the numerous complex relationships in the real-world scenario, compared to the graph.
A number of studies have shown that HGNNs outperform GNNs in many deep learning tasks, such as graph visualization~\cite{huang2009video,zhao2018learning}, bio-informatics and medical image analysis~\cite{tian2009hypergraph,gao2015mci} and recommendation system~\cite{fang2014topic}.

\textbf{
However, there is little research on adversarial attacks against HGNNs, although similar work on GNNs has been extensively studied.}
GNNs are easily fooled by malicious adversarial examples (AEs) and output incorrect results~\cite{SzegeInt13}. 
Recent studies show that HGNNs outperform GNNs in many deep-learning tasks, but they neglect the safety of HGNNs.
\textbf{Moreover, the efficiency of adversarial attacks is severely limited though existing adversarial attack methods of GNNs can be adapted to HGNNs.}
For example, when we exploit the integrated gradient~\cite{wu2019adversarial}, which is used for structure attack on GNN, the time cost of a successful attack on a single node is 40 seconds on average (more detail of our motivation is given in Section~\ref{sec2}).
Since the Cora dataset has 2000 nodes in the hypergraph, the time cost of attacking all the hypergraph nodes is unacceptable. 
Even worse, The problem of limited attack performance is more severe for hypergraphs built on real-world data that have more nodes on them
.
Therefore, despite the superior performance of HGNN, it is still challenging to guarantee its safety.

In this paper, we propose the first adversarial attack framework against HGNNs, called HyperAttack, which focuses on white-box structure attacks against HGNNs.
The goal of HyperAttack is to mislead HGNN and output incorrect node classification results by modifying the hypergraph structure.
For each attacked node in the hypergraph, HyperAttack disturbs the connectivity state of its hyperedges, i.e., adds or removes certain hyperedges of the target node.
The perturbation priority of hyperedges is determined by the guidance of calculated gradients and integrated gradients to improve attack efficiency.
To evaluate the efficiency and effectiveness of HyperAttack, we first consider three different approaches to generating hypergraphs and pre-training HGNN models using popular datasets, i.e., Cora and PubMed, respectively.
We then perform white-box structure attacks on these HGNN models and evaluate the performance of the attacks.
Compared to white-box structural attack methods for GNNs, which use the fast-gradient algorithm (FGA)~\cite{chen2018fast} and the integrated-gradient algorithm (IGA)~\cite{wu2019adversarial}, HyperAttack achieves a 10–13X improvement in time efficiency for successful attacks while also increasing attack success rates by 1.3$\%$-1.7$\%$.
Our main contributions are:

\begin{itemize}
\item  We present HyperAttack, a framework for white-box structural attacks designed for hypergraph neural networks. To the best of our knowledge, this is the first work on adversarial attacks against HGNNs.
\item We clarify the differences between GNN and HGNN structure attacks, analyze the difficulties of simply applying GNN structure attack methods to HGNNs, and present the main challenges of HGNN structure attacks.
\item We propose a multi-gradient guided hyperedge selection algorithm to determine hyperedge perturbation priorities, and the experimental results demonstrate that the method can improve the performance of structure attacks for HGNNs.
\end{itemize}


\section{Background and motivation}\label{sec2}

In this section, we first briefly introduce the network structures of GNN and HGNN.
Then, we show the existing structure attack methods of GNN and clarify the difficulties of applying them to HGNNs.
Finally, we conclude the main challenges of HGNN structure attacks.

\subsection{Structure of GNN and HGNN}

Though Graph Neural Networks (GNN) and Hypergraph Neural Networks (HGNN) are designed to derive knowledge embedding by aggregating the relationship among data, their network structure and required inputs are different.
We first introduce the structure of GNN and HGNN, respectively.
Then, we introduce the main difference between GNN and HGNN applications.

\textbf{GNN and graph modeling.}
All GNNs use data features and relationships between data as inputs, and the relationships are formulated by graph modeling.
The graph modeling is introduced as follows.
Given a graph as $\bm{G} = (\bm{V}, \bm{E})$, where $\bm{V}$ is the nodes sets and $\bm{E}$ is the edges sets.
$\bm{X}\in\mathcal{R}^{|V|\times|D|} $ is the feature of each node.
$\bm{D}$ represents the dimension of the node feature.
$\bm{A}\in\mathcal{R}^{|V|\times|V|} $ is the adjacency matrix, which 
represents the relationship between each node.
If the pair of nodes $\bm{v_i}$ and $\bm{v_j}$ are connected, we set the corresponding element $\bm{A_{ij}}$ in $\bm{A}$ to 1, otherwise set it to 0.
After we model the graph structure by using the node feature matrix $\bm{X}$ and the adjacency matrix $\bm{A}$, we input $\bm{X}$ and $\bm{A}$ into the propagation module~\cite{KipfW16GCN}, and each layer can be formulated as follows: 
\begin{equation}
X^{l+1}=\sigma(\widetilde{D}^{-1/2}\tilde(A)\widetilde{D}^{-1/2}H^lW^l)
\label{eq:1}
\end{equation}
where $\bm{X^{l}}$ and $\bm{X^{l+1}}$ represent the input and the output of the layer, respectively.
$\bm{W}$ is the weight matrix of the $\bm{l}-th$ layer, $\sigma$ indicates the nonlinear activation function.
$\tilde(A)$ represents the adjacency matrix with self connection, and $widetilde{D}$ is the degree matrix of the self connected matrix.

The aggregated information can be captured through the convolution operator to propagate information between nodes.


\textbf{HGNN and hypergraph modeling.}
Hypergraph is a kind of graph type with more complex modeling.
It could provide more information on nodes and their connections via a hyperedge.

Generally, the HGNN model can be divided into hypergraph modeling and hypergraph learning~\cite{gao2020hypergraph}.  
Here, we introduce the hypergraph modeling and lay the relevant contents of hypergraph learning later in Section~\ref{sec7}.

A hypergraph can be denoted as $\mathcal{G}$, which consists of a set of nodes $\mathcal{V}$ and a set of hyperedges $\mathcal{E}$.
We concatenate the hyperedge groups to generate the incidence matrix $\mathcal{H}\in\mathcal{R}^{|\mathcal V|\times\|\mathcal{E}|}$, which represents the relationship between each node and hyperedges.
Each entry $\mathcal{{H}(\mathcal{v},\mathcal{e})}$ indicates whether the node $\mathcal{v}$ is in the hyperedge $\mathcal{e}$.
\begin{equation}\label{eq:2}
H_{ve}=\left\{
\begin{aligned}
1 & , & if \quad v\in e, \\
0 & , & if \quad v\notin e.
\end{aligned}
\right.
\end{equation}
After constructing the hypergraph, $\mathcal{H}$ and the node feature matrix $\bm{X}$ are fed into the pre-trained HGNN to predict the labels for the unlabelled nodes.
Hyperedge convolution~\cite{feng2019hgnn} can be formulated as follows:
 
\begin{equation}\label{eq:3}
X^{l+1}=\sigma(D_v^{-1/2}HWD_e^{-1}\mathcal{H}^TD_v^{-1/2}X^{(l)}\Theta^{(l)}))
\end{equation}
where  $X^{(l)}$ is the signal of hypergraph at $\bm{l}$ layer, $W$ represents the weight of each hyperedge, and $\bm{D_e}$, $\bm{D_v}$ represent the diagonal matrices of the hyperedge degrees and the node degrees, respectively.
$\mathcal{H^T}$ represents the transposed matrix of $\mathcal{H}$, 
$X^{(0)}=X$ and $\sigma()$ denotes the nonlinear activation function, and $\Theta$represents the parameter to be learned during the training process.


\begin{figure}[t]
\centering
\includegraphics[width=1\linewidth]{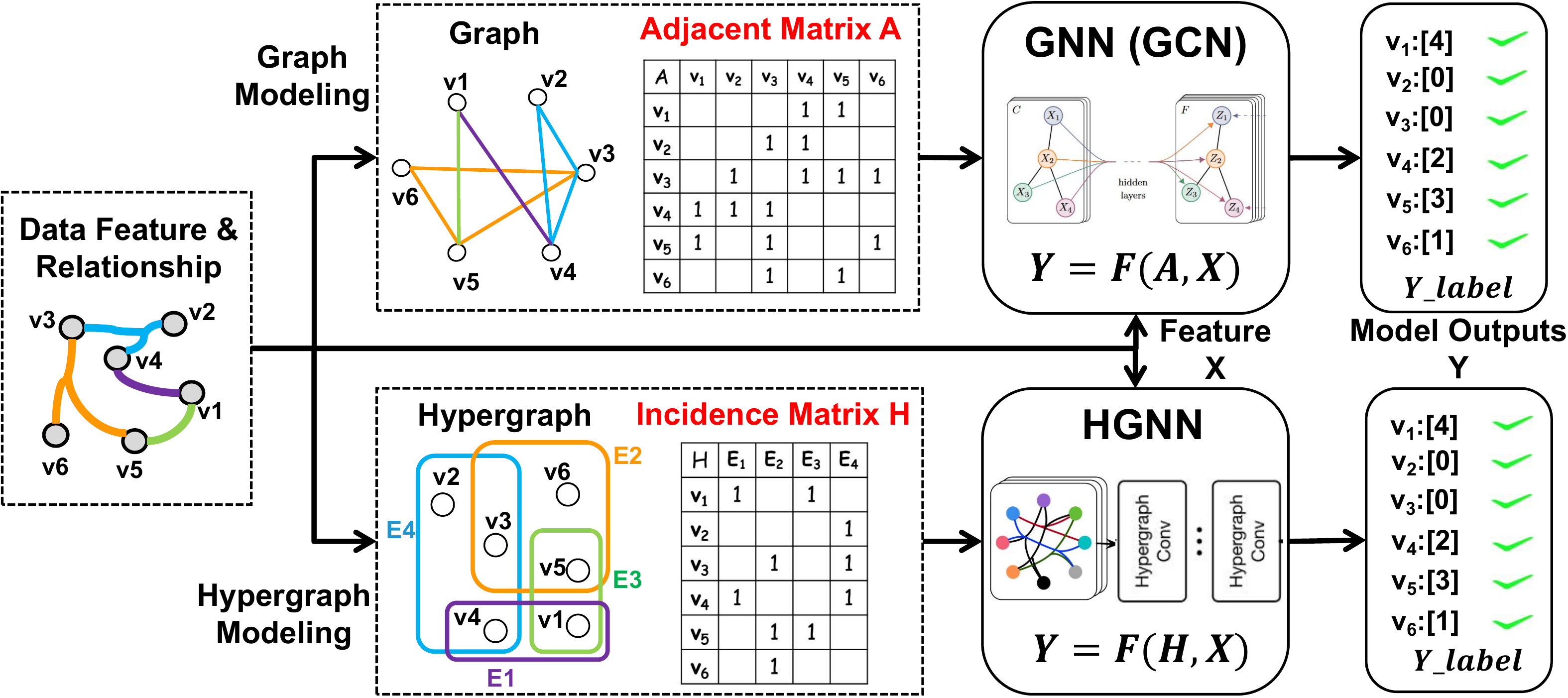}
\caption{Schematic representation of the difference between GNN and HGNN.}
\label{fig_difference1}
\vspace{-0.5cm}
\end{figure}

\textbf{Main differences between HGNN and GNN.}
Figure~\ref{fig_difference1} describes the schematic representation of getting the classification results through GNN and HGNN.
As shown in the figure, the GNN and HGNN include two main stages: graph modeling and graph learning. 
In the upper part of the figure, GNN keeps the pairwise relevance of nodes from the original graph data into the adjacency matrix $\bm{A}$.
In the lower part, HGNN keeps the relationship of nodes and hyperedges and constructs the incidence matrix $\mathcal{H}$, which involves an uncertain number of hyperedges connecting the nodes of the same class. 
Together with the feature matrix $\bm{X}$, both $\bm{A}$ and $\mathcal{H}$ make up the input of the forward propagation in graph and hypergraph learning stages, respectively.
The classification performance can be determined by comparing the output with the real node label.

There is a huge difference in terms of dimension and the internal correlation meaning.
To be specific, $\mathcal{H}$ concentrates on which nodes should be included in each hyperedge, while $\bm{A}$ focuses on the node relationships.
We believe that a major difference between HGNN and GNN remains in the graph modeling process with the different construction strategies between the adjacency matrix and the incidence matrix.


\subsection{Structure Attack of GNN and HGNN}

Although both structure attacks are used to mislead the model by changing the edge connectivity relationships, the difference in relationship modeling, network structure, and inputs between GNN and HGNN lead to differences in the idea of structure attacks on the two models.
Here, we introduce the structure attack of GNN and HGNN, respectively.
Then, we analyze the difference between structure Attacks on GNN and HGNN and the Difficulties of Applying GNNs Structure Attack Methods to HGNNs.

\textbf{Structure Attack of GNN.}
One of the most common strategies of the GNN structure attack is deleting or adding specific edges on the original graph.
It is a key problem to know which edges have the greatest impact on the classification ability of the target model.
In white-box settings, gradient information is often used as the basis for judging the priority of each edge.
For example, some works use gradients as indicators.
They select the pair of nodes with maximum absolute gradient~\cite{chen2018fast} or reflect a better effect on perturbing certain edges indicated from the integrated gradient~\cite{wu2019adversarial}.
Some other works, such as~\cite{Yi2019Nodeinject,sun2020non}, attempt to inject fake nodes and link them with some benign nodes in the original graph, which increases the classification error rate without changing the primary structure between existing nodes.
However, they focus too much attention on the attack success rate, which has the potential to be time-consuming.

\textbf{Structure Attack of HGNN.}
In view of the inherent attribute of preserving the data relevance of hypergraphs, we can still learn from existing attack methods to address structure disturbance.
One of the simplest strategies is to randomly perturb the inherent structure of the hypergraph.
For example, inspired by~\cite{waniek2018hiding}, we randomly connect nodes with different labels with newly generated hyperedges, or discard specific existing hyperedges between nodes sharing the same label.
Besides, using gradients as information to guide structure attacks can produce more effective changes on a hypergraph.
We can use two typical algorithms, Fast Gradient Sign Method (FGSM)~\cite{goodfellow2014explaining} and Jacobian-based Saliency Map Approach (JSMA)~\cite{papernot2016limitations}, which shows the success for image data and be studied firstly by~\cite{wu2019adversarial} for graph models.

\begin{figure}[t]
\centering
\includegraphics[width=1\linewidth]{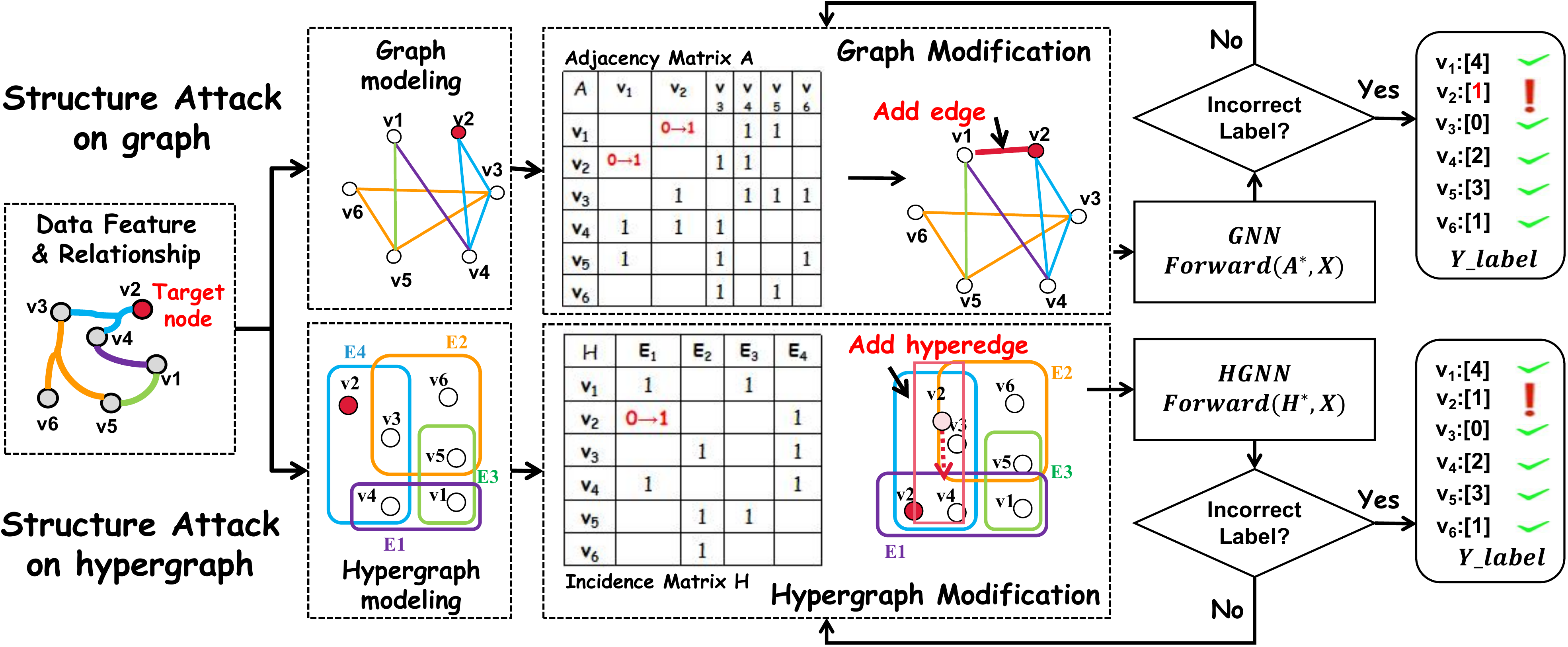}
\caption{Difference of structure attack against hypergraph and graph.
(a) Given a node set $(v_1,...v_6)$, we attack the target node ${v_2}$ in the pre-trained GNN and HGNN models.
(b) In the upper part, by adding a specific edge connecting the target node ${v_2}$ with ${v_1}$, the adjacency matrix structure is modified. 
The attack ends when the classification of target node $\bm{v_2}$ changes.
(c) In the lower part, we modify the relationship between $\bm{v_2}$ and the hyperedge $\bm{E1}$ in $\mathcal{H}$, which means we have added the target node to a new hyperedge.
At last, we input the modified $\mathcal{H}$ into the forward propagation of HGNN to judge the classification of the target node.
}
\label{fig:difference}
\vspace{-0.3cm}
\end{figure}

\textbf{Difficulties of Applying GNNs Structure Attack Methods to HGNNs.}
\begin{itemize}

\item The structure attack on HGNN is realized by modifying the internal information of the incidence matrix.
Different from modifying the adjacency matrix of GNN, this unique modification is currently in the blank, which brings the first difficulty.

\item Because the construction of HGNN is more complex in preserving data correlations compared with GNN, directly migrating existing graph attack methods has been proven in experiments for causing unsatisfactory performance, especially in terms of time complexity.
\end{itemize}

Through some direct migration experiments (e.g., integrated-gradient-based attack~\cite{wu2019adversarial}).
For example, it takes 40 seconds at least to fully implement the attack on one single target node When we set $2$ $*$ Tesla V100S GPU as our computation resources. 
Imagine the implementation against the entire dataset (e.g., the Cora dataset contains 2708 scientific publications as nodes), the time cost will be more than 30 hours.
Worse still, this is only the cost when we assume that all the attacks succeed.
Nevertheless, in the case of worse computing resources, the time burden will be more serious.

As a result, a large gap between the increasing-mature GNN testing and deficient HGNN testing methods still exists.




\subsection{Challenges of HGNN structure attacks}
Based on the clarified difference between HGNN and GNN, as well as the difficulties of Applying GNNs Structure, we introduce the main challenges of HGNN structure attacks.

\textbf{Lack of modification on hypergraph structure.}
Compared with graphs, the dimension of hypergraph structure is even larger in many cases.
The dimension is directly proportional to the complexity in the face of some complex origin data, such as multi-modal data or data with a high-order association.
It will spend more overhead to obtain the internal information and address structure attack on the HGNN rather than GNN.
For example, when determining a specific node, traversing and computing on the incidence matrix $\mathcal{H}$ may cause more burden, which leads to the existing methods of modifying the structure matrix can not be well applied.
The modification of the hypergraph structure still lacks practice strategies.

\textbf{Balance between attack success rate and time consumption.}
A superior attack algorithm should balance the time cost and attack success rate.
However, graph attack algorithms in the past paid too much attention to the attack success rate in a long term.
In the premise of a successful attack, reducing the running time of generating adversarial samples should not be ignored.
Therefore, it is another challenge for us to realize HyperAttack with both a high attack success rate and low time cost.

\section{Threat Model}\label{sec3}
The goal of our structure attack is to reduce the classification ability of the target node.
We expect to attack the target node of a hypergraph by changing the structure within the threshold of a constraint.
In this section, we clarify the structure attack scenario of HyperAttack by introducing the setting target model, target objects, and specific attack actions. 

\subsection{Target Model}
We consider a two-layer HGNN model~\cite{feng2019hgnn} which is a general framework for hypergraph representation learning as our target model.
Based on Section~\ref{sec2}, the description of the target model can be stated as:

Given $\mathcal{G}=(\mathcal{H},X) $ with a set of labeled nodes ${v_t}$ where
$\bm{Y_l}$ is the ground truth of each node.
$\mathcal{H}$ and $\bm{X}$ represent the incidence matrix and node feature matrix, respectively.
Our target model focuses on the node classification task to categorize nodes without labels into different classes by using a large number of labeled nodes.
After the training process as mentioned in Eq~\ref{eq:4}, the target model can complete the semi-supervised learning task and predict the classification result.
We set the target node as ${v_t}$, and attack ${v_t}$.
In the classification results, the target model will be misled only when facing the target node.

\subsection{Structure Attack Objectives}
Because the HGNN model first constructs hypergraph from original data, the strategies of generating hypergraph directly affect both the quality of the prediction ability and the stability under adversarial attack.
The hypergraph generation methods vary a lot and we introduce the details in Section~\ref{sec7}. 
Here, we consider the incidence matrix of the hypergraph as the objective in our structure attack and introduce three classic strategies to verify the different results of different construction strategies under the same attack algorithm.

Based on K nearest neighbors (KNN) methods, we follow~\cite{huang2009video} and construct hyperedges by connecting each node with their nearest $K$ nodes, called Hypergraph\_KNN.
Based on $\epsilon$-ball, we use another distance-based method to construct hyperedges by connecting nodes whose distances are less than the preset threshold, called Hypergraph\_$\epsilon$.
Based on the $\bm{L1}$ reconstruction method, we followed~\cite{WangLW15} to generate each hyperedge via modeling the relationship between nodes through feature reconstruction, called Hypergraph\_L1.
We also show the different robustness of the above three modeling strategies under attacks in Section~\ref{sec7}.

\subsection{Structure Attack Operations}
In this subsection, in order to make our experiments more interpretable, we introduce some attack actions behind the implementation of our proposed structure attack.
Our goal is to find out which elements in $\mathcal{H}$ have the maximum impact on the classification results after being attacked.
We further clarify our attack actions as deleting (or adding) the target node on the selected hyperedges, which modifies the incidence matrix $\mathcal{H}$ indeed.
As a highly directional attack, modifying the relevant relationship of the target node has little tiny impact on other nodes.
This makes HyperAttack only change the classification ability of the target node, which well shortens the attacking burden on a small budget.

It is worth mentioning that we abandon another attack action in the experiment:  creating a new hyperedge with the target node.
Because the newly created hyperedge will change the dimension of the original matrix $\mathcal{H}$, which may be a huge burden to HyperAttack. 
Worse still, other nodes placed in the new hyperedge will also bring potential damages to the global classification ability of hypergraph.

\section{Methodology}\label{sec4}
In this section, we introduce the framework and algorithm of HyperAttack.
For convenience, we briefly summarize some important symbols and corresponding definitions in table1~\ref{table:notation}. 

\begin{table}[]
\centering
\small
\caption{\label{table:notation} Notation And Definition }
\begin{tabular}{cc}
\hline
Symbol & Definition \\ 
\midrule
$\bm{G=(V,E)}$                                            & input graph $\bm{G}$ with nodes $\bm{V}$ and edges $\bm{E}$\\
$\bm{\mathcal{G}=(\mathcal{V},\mathcal{E})}$              & input hypergraph $\mathcal{G}$ with nodes $\mathcal{V}$ and hyperedges $\mathcal{E}$\\
$\mathcal{G^{*}}$                                         & output hypergraph $\mathcal{G^{*}}$ \\
$\bm{D_v}$                     & the diagonal matrices of the node degrees  \\
$\bm{D_e}$                     & the diagonal matrices of the edge degrees  \\
$\bm{W}$                       & the diagonal matrix of the hyperedge weights  \\
$\bm{A}$                       & the adjacency matrix of GNN  \\
$\mathcal{H}$                  & the incidence matrix of HGNN  \\
$\bm{X}$                       & the node feature matrix   \\
$\bm{D}$                       & the dimension of the node feature\\
$\mathcal{\Bar{H}}$            & the convolved signal matrix of $\mathcal{H}$   \\
$\bm{{F}}$                     & the forward propagation output of the HGNN \\  
$\bm{Y_l}$                     & the real label confidence list          \\ 
${v_t}$                        & the target node          \\
$M$                            & the size of fast-gradient-based hyperedge filter  \\ 
$N$                            & the size of integrated-gradient-based hyperedge selection \\
$\mathcal{H^{*}}$              & modified incidence matrix \\
$\gamma$                       & number of disturbance limit of ASR \\
$\eta$                         & number of disturbance limit of AML \\
\bottomrule
\end{tabular}
\end{table}

\begin{figure}[ht]
\centering 
\includegraphics[width=1\linewidth]{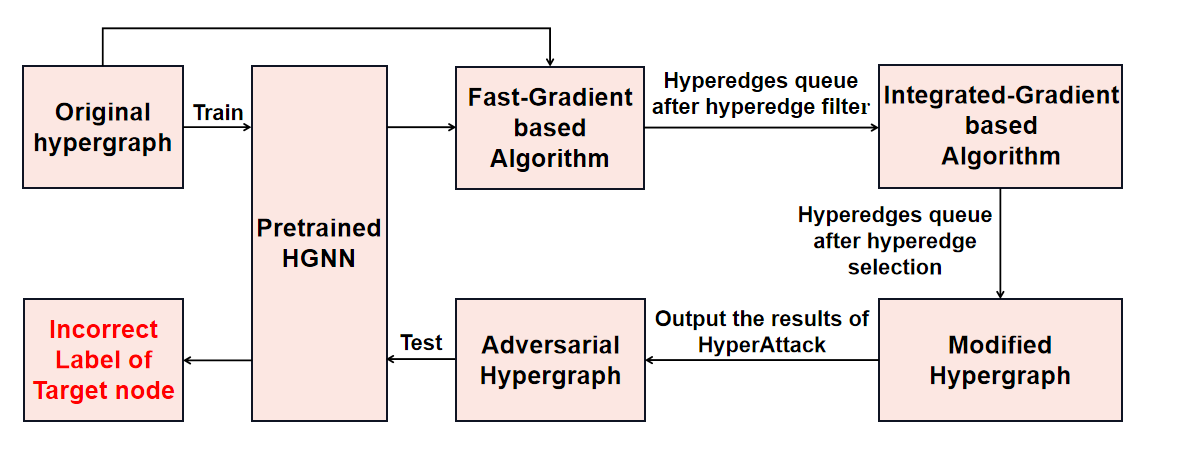}
\caption{\label{fig_framework} The framework of HyperAttack.
}
\vspace{-0.5cm} 
\end{figure}

\subsection{Overview}


As shown in Figure~\ref{fig_framework}, there are three main parts in HyperAttack, the Fast-Gradient based algorithm part, the Integrated-Gradient based algorithm part, and the modified hypergraph part, out of which the selection process is made up of the first two parts, and the modification process is the third part.

In comprehensive consideration of the factors with the lowest time cost and high attack success rate,
HyperAttack is well-designed through a fast-gradient-based hyperedge filter using gradients and integrated-gradient-based hyperedge selection using integrated gradients.
Both gradients and integrated gradients are the indicators to evaluate the priority of each hyperedge to the target node.

To be specific, we first obtain the gradient of the whole $\mathcal{H}$ by calculating a designed loss function.
Then we keep the hyperedges with the maximum absolute value of gradient as the result of a fast-gradient-based hyperedge filter.
Secondly, by further setting another loss function, the integrated gradients of each hyperedge from the fast-gradient-based hyperedge filter are obtained.
The hyperedges with maximum value will be selected as the result of fine-grained screening.
In the third step, we regard the hyperedges from fine-grained screening as the final selection.
We remove or add the target node on these hyperedges via modifying the incidence matrix $\mathcal{H}$.
The details are as follows.

\subsection{Fast-gradient-based hyperedge filter}

As mentioned in the target model, we use a two-layer HGNN model~\cite{feng2019hgnn} for the node classification task with an incidence matrix $\mathcal{H}$ and a node feature matrix $\bm{X}$ as the input of the forward propagation.
As mentioned in Eq~\ref{eq:4}, we use $\mathcal{H}$ as the variable of the output prediction $\bm{F}$($\mathcal{H}$).
The HGNN model forward propagation obtains the output $\bm{F}$ of the last layer, which can be simply described as $\bm{F}=forward(\mathcal{H},\bm{X})$ and its simple form are as followed:
\begin{equation}\label{eq:4}
{\bm{F}}(\mathcal{H})=f(\left(\mathcal{\Bar{H}}\sigma\left(\mathcal{\Bar{H}}\bm{X}\mathcal{W_0}\right)\mathcal{W_1}\right)
\end{equation}
where $\mathcal{\Bar{H}}$ is formulated as shown in Eq~\ref{eq:5}.
$\bm{W_0}$ and $\bm{W_1}$ are the input-to-hidden and hidden-to-output weight matrices,respectively.
$f$ represents the $softmax$ function and $\sigma$ is the $Relu$ active function. 
It should be noted that $\mathcal{H}$ is the only variable of forward propagation. 
HyperAttack does not add disturbance on $\bm{X}$, thus $\bm{X}$ remains unchangeable and can be regarded as a constant as the same as $\mathcal{}{W_0}$ and $\mathcal{W_1}$.
\begin{equation}\label{eq:5}
\bm{\Bar{\mathcal{H}}}=D_v^{-1/2}HWD_e^{-1}\mathcal{H}^TD_v^{-1/2}
\end{equation}

We take the difference between the output prediction $\bm{F}$ and the real result $\bm{Y_l}$ and use a simple cross-entropy to express the discrepancy.
Aiming at the target node $v_{t}$, we formulate our designed function $L_{t}$ as:
\begin{equation}\label{eq:6}
L_{t}=-\sum_{k=0}^{|\mathcal{E}|-1} {\bm{Y_{ltk}}ln\left({F}_{tk}\left(\mathcal{H}\right)\right)}
\end{equation}

We calculate the partial derivatives in Eq~\ref{eq:6} with respect to $\mathcal{H}$ and get $g_{tk}$ as the $k^{th}$ gradient of the target node $\mathcal{v_{tk}}$.
\begin{equation}\label{eq:7}
g_{tk}=\frac{\partial L_{tk}}{\partial \mathcal{H}_{tk}}
\end{equation}

We set a hyper parameter $\bm{M}(0<M<|E|)$.
The hyperedges within top-$M$ largest gradients will be recorded as the results of the Fast-gradient-based hyperedge filter.
The pseudo-code for the fast-gradient-based hyperedge filter is given in lines 3 to 4 of Algorithm 1.

\subsection{Integrated-gradient-based hyperedge selection}

The results of preliminary filtering using fast gradient-based methods have been saved.
Limited by the number of perturbations, we are inspired by ~\cite{sundararajan2017axiomatic,wu2019adversarial} and we choose the integrated gradient-based method to realize hyperedges selection.
The integrated gradients-based method combines direct gradient and back-propagation-based approaches.
Let $x$ be the input value and $x'$ be the baseline value.
The function mapping is expressed as $F$.
The Integrated gradient of the $i^{th}$  input can be expressed as follows:
\begin{equation}
IntergratedGrads_{i}\left ( x \right ) : : = \left(x_i - x_i^{'}\right) \times \int_{\alpha = 0}^{1}{\frac{\partial F\left(x' +\alpha \times \left(x_i - x_i^{'}\right)\right)}{\partial x_i}} d\alpha
\end{equation}\label{eq:8}

Since the gradient of all points on the whole path is considered, it is no longer limited by the gradient of a specific point.
For target node ${v_t}$,we set the matrix $H_a^{'}$ to all-one and the matrix $H_r^{'}$ to all-zero matrix respectively.
The matrix $H_a^{'}$ and matrix $H_r^{'}$ represent the target note $v_t$ with all hyperedges connected and fully unconnected, respectively.  

\begin{equation}\label{eq:9}
H^{'}=\left\{
\begin{aligned}
H_a^{'} : H\left[t\right]\left[i\right] = 1 , 0 \le i < \left | E \right |  \\
H_r^{'} : H\left[t\right]\left[i\right] = 0, 0 \le i < \left | E \right | 
\end{aligned}
\right.
\end{equation}

For target node $v_t$,  when there is no connection between the target node $v_t$ and the hyperedge $i$, we set the $H_a^{'}$ matrix as a baseline since we want to describe the overall change pattern of the target function $F$ while gradually disconnect the target node $v_t$ from the hyperedges to the current state of $H$. 
On the contrary, when the target node $v_t$ is already connected to the hyperedge $i$, we use the $H_r^{'}$ matrix as a baseline to calculate the changing pattern by gradually adding the connection between $v_t$ and the hyperedges.

\begin{equation}\label{eq:10}
\begin{split}
&IG\left(F\left(H,t\right)\right)\left[t,i\right]=      \\
&\left\{
\begin{aligned}
\left(H_{ti} - 0\right) \times \sum_{k=1}^{m}{\frac{\partial F\left({H_r^{'}} +\frac{k}{m}\left(H- H_r^i\right)\right)}{\partial H_{ti}}} \times \frac{1}{m},H\left[t\right]\left[\i\right] \ne 0 , \\
\left(1 - H_{ti}\right) \times \sum_{k=1}^{m}{\frac{\partial F\left({H_a^{'}} -\frac{k}{m}\left(H_a^{'}- H\right)\right)}{\partial H_{ti}}} \times \frac{1}{m},H\left[t\right]\left[\i\right] = 0.
\end{aligned}
\right.
\end{split}
\end{equation}

Lines 5 to 11 of Algorithm 1 show the pseudo-code for the Integrated-gradient-based hyperedge selection of HyperAttack. 
We calculate the integrated gradient of the hyperedges and then save $N$ hyperedges with the largest gradient, where $N$ is the size of perturbations.

\subsection{Matrix-modified operation}

The result of HyperAttack is an adversarial hypergraph network with a modified matrix $\mathcal{H^{*}}$ which can ensure the original HGNN model produces the wrong classification for the target node in forward propagation.
In consideration of the notion of ’unnoticeable changes’ in hypergraph, we set a strict upper limit on the number of perturbations.
We use the selected index of the hyperedges from the fine-grained screening to modify the links of the original incidence matrix $H$ which is defined as:
\begin{equation}\label{eq:10}
\mathcal{H^{*}}= \mathcal{H} + \theta\left(\bm{g_i}\right),
\end{equation}
where $\theta\left(g_i\right)$ represent the sign of adding/removing the $i^{th}$relationship of the target node and links. 
The overall HyperAttack algorithm is summarized by Algorithm 1.

\begin{figure*}
\centering 
\includegraphics[width=1\linewidth]{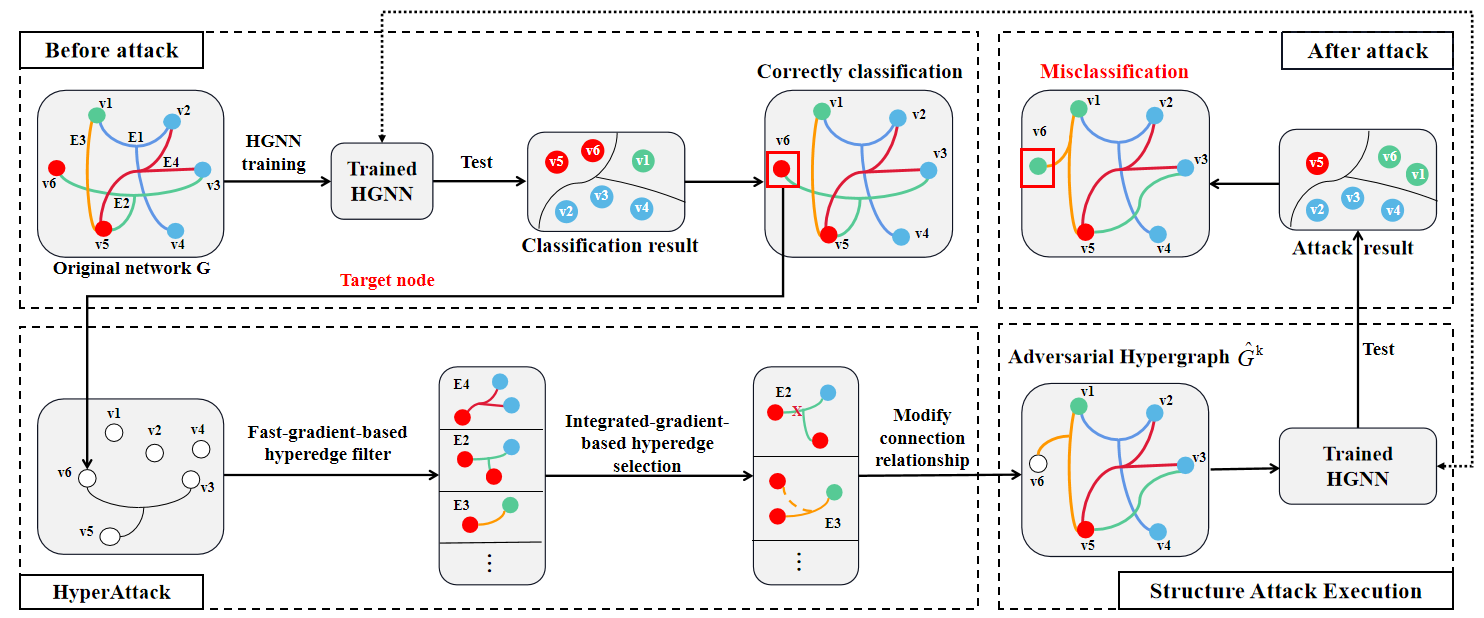}
\caption{The flowchart of HyperAttack.}
\vspace{-0.2cm} 
\label{fig:flowchart}
\end{figure*}

\begin{algorithm}[t]
\SetAlgoLined
\normalem
\SetKwInOut{Input}{input}\SetKwInOut{Output}{output}
\caption{Multi-Gradient-Guided Structure Attack of HGNN}
\label{alg:1}
\KwIn{ Original incidence matrix $\mathcal{H}$, 
number of coarse-grained screening $\bm{M}$,
the maximum number of perturbations $\bm{N}$,
target node $v_t$\;} 
\KwOut{Modified incidence matrix $\mathcal{H^*}$\;}
Initialize $\kappa = 0$, $i = 0$\;
Train the HGNN model on the original network $H$.\\
// Obtain the gradient of the whole $H$ via Eq~\ref{eq:6}, Eq~\ref{eq:7} .\\
coarse-grained screening $\leftarrow$ Index of the top-M hyperedges with the largest gradient\;
\While{$\kappa < M$}{
    //Obtain the $\kappa$th hyperedges as the coarse-grained screening.\\ 
    $i \leftarrow$ coarse-grained screening$[\kappa]$\;
    // Obtain the integrated gradient of the hyperedge i via Eq~\ref{eq:9}.\\
    $g_{i} \leftarrow IG[v_t,i]$\;
    $\kappa++$\;
}
fine-grained screening $\leftarrow $ Select the Index of the $N$ hyperedges with the the largest integrated gradient\\
// Modify the matrix $H$ according to the fine-grained screening.\\ 
$H^* \leftarrow H + \theta\left(g_i\right)$\; 
\Return $\bm{H^*}$\;
\end{algorithm}

\section{Evaluation}\label{sec5}

\subsection{Evaluation Design} 
\subsubsection{Research questions}
We conduct experiments and try to answer the questions as follows.



\textbf{RQ1. How is the performance of HyperAttack in attack
 success rate compared to other state-of-the-art methods?}
In this RQ, we compare the attack success rate of HyperAttack with some state-of-the-art attack methods on graph.
We wonder whether these graph attack methods can be adopted on HGNN with a high attack success rate.

\textbf{RQ2. How is the performance of HyperAttack in time
cost when completing a successful attack compared to other state-of-the-art methods?}
In this RQ, we wonder about the efficiency of HyperAttack and the others.
We use running time to record the time consumption required for a successful attack using HyperAttack and other methods.
We consider the attack method which has higher efficiency and is able to spend less running time while achieving the same attack effect.

\textbf{RQ3. How does the classification margin perform under HyperAttack compared to other state-of-the-art methods?}
In this RQ, we calculate the classification margin for all methods and analyze the characteristics of HyperAttack.

\subsubsection{Evaluation environment}
Our experimental environment consists of 2$*$Intel(R) Xeon Gold 6248 CPU @2.50GHz$\times 40$, 2$*$Tesla V100S GPU, 384 memory, and CentOS 7.
The max number of perturbations $N$ is 10.
We use three different hypergraph generation approaches to formulate the data correlation from the original data, which consists of the K-Nearest Neighbor-based (KNN) method,  $\varepsilon$-ball-based method, and $l1$-hypergraph-based method.
The reason for using different hypergraph modeling is that the sensitivity of HGNNs to adversarial attacks is different, and the robustness of the attack performance needs to be evaluated.

\subsubsection{Evaluation metrics}
As we mentioned in Section \ref{sec2}, we stated that a successful structure attack must have two key characteristics:
(1) achieving the desired attack performance;
(2) balancing the attack time consumption.
Therefore, we use the Attack Success Rate (ASR) and Average Number of Modified Links (AML) to evaluate structure attack performance. 
Then, Running Time (RT) is used to evaluate the time cost of the structure attack.

\begin{itemize}
\item{Attack Success Rate (ASR).}
Given a hyper-parameter $\gamma (0<\gamma \le 10)$. 
Each target node changes no more than $\gamma$ links.
ASR represents the success rate of each target node under a certain number of attacks.
The formal expression is as follows:
\begin{equation}
ASR=\frac{ Successful attacks}{All attacks}    
\end{equation}

\item{Average Number of Modified Links (AML).}
AML calculates the average counts of modifying links for a successful attack.
$\eta$ is used to represent the upper bound of the number of perturbations.
AML describes the average number of modified links the attacker needed to meet the attack objective:
\begin{equation}
AML=\frac{Modified links}{All attacks}    
\end{equation}

\item{Running Time(RT).}
RT records the time consumption required for each successful attack.
It takes the original matrix $\mathcal{H}$ input as the starting time, and the modified matrix $\bm{H^{*}}$ after the successful attack as the ending time, we formulate it as follows:
\begin{equation}
RT =  Time_{end}  -  Time_{start}
\end{equation}

\end{itemize}
\subsubsection{Parameter settings}
We give some parameter settings here which will be used in the experiment.
\textbf{$\gamma$:}
We set the limit number of ASR $\gamma$ to 10 in our experiments, which means the successful attack should be limited to 10 perturbation times, otherwise the attack is considered failed.
In the experiment of calculating the running time, $\gamma$ is set to increase continuously from 1 to 10.

\textbf{$\eta$:}
We set the upper bound of the number of perturbations $\eta$ to 10 in our experiments.
We only consider the successful attack in 10 times and calculate the average number of perturbations.

\textbf{$K$:}
KNN is a traditional method of constructing hyperedges. 
In the experiment part, we fixed the value of $k$ as 10, which means each hyperedge will contain 10 nodes.
In addition, we conduct experiments to verify how $K$ affects the robustness of HGNN in the Discussion part.


\subsubsection{Datasets}
The target models in our experiments are focused on the node classification task.
The node classification task is a semi-supervised learning task that aims to accurately predict the category of each node. 
We have selected two widely-used datasets and some basic statistics for these datasets are provided below.
\begin{itemize}
\item Cora: The Cora dataset consists of 2708 scientific publications classified into one of seven classes.
The citation network consists of 5429 links, which represent the citation relationships.

\item PubMed: The PubMed dataset consists of 19717 scientific publications from the PubMed database pertaining to diabetes classified into one of three classes. 
The citation network contains 88676 links, which represent the citation relationships.
\end{itemize}

\subsubsection{Compared Methods}
Here, we compare HyperAttack with five baseline methods.
In general, we divide the five existing baseline methods into two categories: random-based attack methods and gradient-based attack methods.
We briefly describe them as follows.

\textbf{Random-based Attack.}
In view of the fact that the robustness of graph and hypergraph lack of stability mechanism, the classification performance of the target node will be influenced to a certain extent when the inherent attribute of the structure are changed via random-based methods~(i.e. connect or disconnect edges via modifying original incidence matrix).
In our work, we introduce three random-based methods, including Random Delete~(\textbf{RanD}), Random Modified~(\textbf{RanA}), and Disconnect Internally, Connect Externally~(\textbf{DiceA})~\cite{waniek2018hiding}.
Particularly, given $\bm{c}$ and $\bm{C}$ as the actual disturbance times and the maximum limit of disturbance times, RanD randomly selects $c (c < C) $ hyperedges from the hyperedges connected to the target node and then disconnect  the target node from the selected hyperedges.
RanA randomly selects $c (c < C) $ hyperedges and can disconnect or connect the relationship between the target node and the selected hyperedges.
DiceA first randomly disconnects $\frac{c}{2} (c < C)$ hyperedges of the target node, then randomly connect the target node on $|E|-c$ non-selected hyperedges.

\textbf{Gradient-based Attack.}
The gradient-based attack is widely used as an efficient algorithm in many types of research, especially in methods based on white-box settings.
When the internal information of the target model is available for attackers, the simplest approach is to use the gradient of the training process as indicators to evaluate the priority of each component between nodes and hyperedges.
Attackers can then modify the matrix according to the priority.
In our work, we introduce two gradient-base methods, including \textbf{FGA}~\cite{chen2018fast} and \textbf{IGA}~\cite{wu2019adversarial}, which use gradient and integrated gradient, respectively.

\subsection{Structure Attack Effectiveness}

\begin{table*}[t]
\renewcommand\arraystretch{1.2}
\setlength\tabcolsep{2pt}
\centering
\normalsize
\caption{\label{table:2} Structure attack performance (ASR, AML, and RT) of five structure attack methods: RanD, RanA, DiceA, FGA, IGA, and HyperAttack with different hypergraph models.
The performance of HyperAttack shows the best results in both attack success rate and time cost.}
\begin{tabular}{c|cc|ccccccccc}
\hline
\multirow{2}{*}{Dataset} & \multicolumn{2}{c|}{\multirow{2}{*}{Method}}                                                                      & \multicolumn{3}{c|}{Hypergraph\_KNN} & \multicolumn{3}{c|}{ Hypergraph\_$\epsilon$} & \multicolumn{3}{c}{Hypergraph\_L1} \\ \cline{4-12} 
                         & \multicolumn{2}{c|}{}                                                                                              & ASR(\%)     & AML      & RT(s)     & ASR(\%)      & AML      & RT(s)     & ASR(\%)         & AML          & RT(s)         \\ \hline
\multirow{6}{*}{Cora}    & \multicolumn{1}{c|}{\multirow{3}{*}{\begin{tabular}[c]{@{}c@{}}Random-based\\ Attack\end{tabular}}}   & RanD         & 15          & 4.20     & 0.02          & 12           & 4.39     & 0.02           & 80              & 1.96         & 0.02               \\
                         & \multicolumn{1}{c|}{}                                                                                 & RanA         & 48          & 4.78     & 0.02           & 52           & 4.34     & 0.02           & 45              & 4.27         & 0.02               \\
                         & \multicolumn{1}{c|}{}                                                                                 & DiceA       & 37          & 5.49     & 0.02           & 34           & 4.88     & 0.02          & 69              & 3.2          & 0.02               \\ \cline{2-12} 
                         & \multicolumn{1}{c|}{\multirow{3}{*}{\begin{tabular}[c]{@{}c@{}}Gradient-based\\ Attack\end{tabular}}} & FGA        & 85          & 3.00     & 0.03           & 91           & 1.37     & 0.04           & 89              & 1.75         & 0.03               \\
                         & \multicolumn{1}{c|}{}                                                                                 & IGA        & 88          & 2.49     & 42.23           & 95           & 1.43     &42.18           & 91              & 1.85         & 44.44               \\ \cline{3-12} 
                         & \multicolumn{1}{c|}{}                                                                                 & HyperAttack (OURS) & 89          & 2.84     & 3.65          & 96           & 1.39     & 3.64           & 96              & 2.02         & 3.74              \\ \hline
\multirow{6}{*}{PubMed}  & \multicolumn{1}{c|}{\multirow{3}{*}{\begin{tabular}[c]{@{}c@{}}Random-based\\ Attack\end{tabular}}}   & RanD         & 11          & 6.45     & 0.08          & 3            & 8.74     & 0.08          & /               & /            & /              \\
                         & \multicolumn{1}{c|}{}                                                                                 & RanA         & 29          & 6.36     & 0.08          & 30           & 6.49     & 0.08          & 40              & 3.92         & 0.08              \\
                         & \multicolumn{1}{c|}{}                                                                                 & DiceA       & 17          & 5.43     & 0.08          & 14           & 7.33     & 0.08          & /               & /            & /              \\ \cline{2-12} 
                         & \multicolumn{1}{c|}{\multirow{3}{*}{\begin{tabular}[c]{@{}c@{}}Gradient-based\\ Attack\end{tabular}}} & FGA        & 48          & 5.22     & 0.16          & 72           & 3.79     & 0.15          & 96              & 1.47         & 0.10              \\
                         & \multicolumn{1}{c|}{}                                                                                 & IGA        & 69          & 5.16     &187.13           & 82           & 3.60     &187.33           & 97              & 1.32         &186.13               \\ \cline{3-12} 
                         & \multicolumn{1}{c|}{}                                                                                 & HyperAttack (OURS) & 74          & 5.11         &19.38           & 83           & 3.68     &19.50           & 97              & 1.26         &19.48               \\ \hline
\end{tabular}
\end{table*}

As shown in Table~\ref{table:2}, we use HyperAttack and five other baseline methods on the Cora and PubMed datasets.
Take the Cora dataset as an example, HyperAttack has great advantages over all Random-based Attack methods where the average ASR has increased from 41$\%$ to more than 92$\%$ when we use the KNN-based method to generate hyperedges.
Compared with the 95$\%$ attack success rate achieved by IGA, which is the state-of-the-art attack method on graph, HyperAttack increased to 96 $\%$.
When we use the $\epsilon-ball$, ASR has increased from 91$\%$ of IGA to 96$\%$ of HyperAttack.
As for the PubMed dataset, ASR has decreased to different degrees under each attack algorithm, but HyperAttack is still far superior to all Random-based methods and shows slightly better than the ASR of IGA.
\textbf{Answer to RQ1:} 
We can preliminarily conclude that HyperAttack has outstanding advantages over the existing Random-based methods, and is slightly superior to IGA by 1.3$\%$ - 3.7$\%$ of ASR.

\begin{figure}[ht]
    \centering
    \subfigure[Cora] {\includegraphics[width=0.49\linewidth]{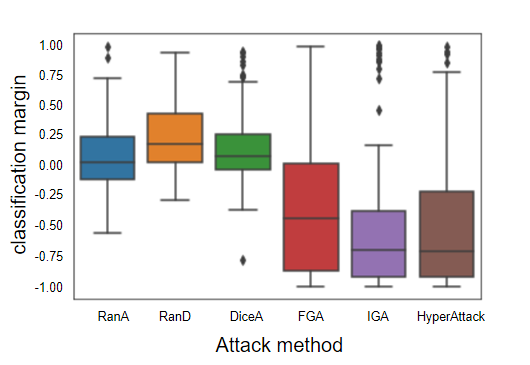}}
    \subfigure[Pubmed] {\includegraphics[width=0.49\linewidth]{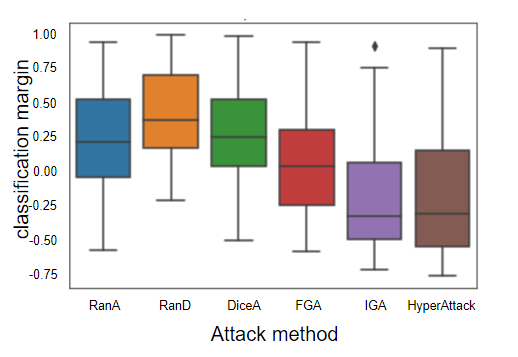}}
    \caption{\label{Classification} {The classification margin of different attack methods on the Cora dataset.}}
    \vspace{-0.2cm} 
\end{figure}

We use classification margin to evaluate the attack performance.
For the target node $v$, its classification margin is $X = Z_{v,c} - max_{c'\neq c}Z_{v,c'}$, where $c$ is the ground truth label, $Z_{v,c}$ is the probability that the target node $v$ is given label $c$ by HGNN model. 
Lower $X$ are better. $X$ smaller than 0 means the target node is misclassified.
Figure\ref{Classification} shows the classification margin of different attack methods on the Cora dataset and PubMed dataset.
Obviously, the classification margin of IGA and HyperAttack is significantly better than other attack methods.
IGA is relatively stable, but HyperAttack has a slightly higher success rate than IGA.
\textbf{Answer to RQ3:} 
HyperAttack has the highest attack success rate with a guaranteed classification margin better than other attack methods except for IGA.
We believe HyperAttack is accurate and relatively stable.

\subsection{Structure Attack Performance}

In addition, we can get the running time of a successful attack process from Table~\ref{table:2}.
We find the running time of all Random-based methods is only 0.02 seconds, because this kind of method only needs the operation time of disturbance, and does not need time to filter from the incidence matrix. 
In contrast, the Gradient-based methods  take longer for they spend most of the time doing the selection.
IGA and HyperAttack both show superior ASR.
However, the time cost of HyperAttack is much shorter than that of IGA.
To be specific, when we implement a successful attack via IGA and HyperAttack respectively, IGA takes 42-45 seconds, while HyperAttack only takes 2-4 seconds, which reduces the time by 10-20 times.
\textbf{Answer to RQ2:} 
We believe that the running time of HyperAttack is 10-20 times shorter than that of similar the-state-of-the-art methods under the condition of ensuring a high success rate of the attack.

\subsection{Visualization and case study}

\begin{figure}[ht]
    \centering
    \subfigure[Before attack] {\includegraphics[width=0.45\linewidth]{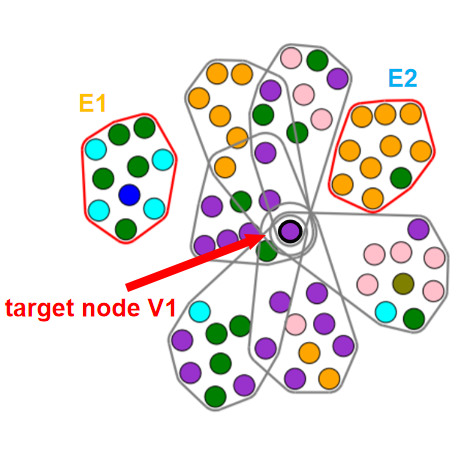}}
    \subfigure[After attack] {\includegraphics[width=0.45\linewidth]{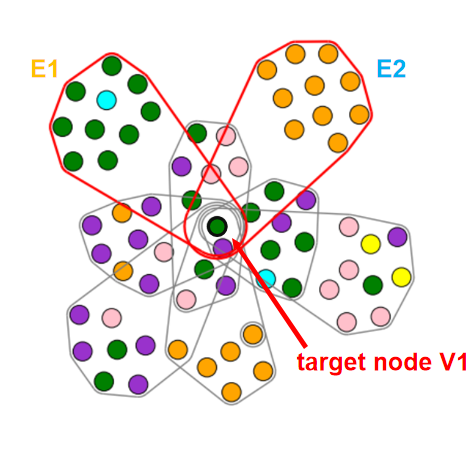}}
    \caption{\label{visualization} {The visualization of HyperAttack. (a) is the connection relationship of the nodes before the attack. (b) is the connection relationship of the nodes after the attack.}}
\vspace{-0.4cm}    
\end{figure}

In order to make HyperAttack easier to understand, Figure~\ref{visualization} shows the visualization of HyperAttack. We visualize the connection relationship of the target node and compare the changes before and after HyperAttack. 
We take a HyperAttack of Hypergraph\_KNN on the Cora dataset as the example, and the perturbation size is set to 2.
After the original hypergraph is trained by HGNN, the connection relationship of the target node $V1$ is shown in Figure~\ref{visualization} (a). 
$V1$ is connected with six hyperedges and belongs to the category represented by purple.
In HyperAttack, we choose hyperedges $E1$ and $E2$ to attack by Fast-gradient-based filter and Integrated-gradient-based selection. 
Then, we modify the incidence matrix and test it in the trained HGNN, the result is shown in Figure~\ref{visualization} (b).
After HyperAttack, the target node $V1$ adds the connection to the hyperedges $E1$ and $E2$.
Its label changes to the category represented by orange which proves the attack is successful.


\section{Discussion}\label{sec6}
In this section, we discuss the robustness of different methods of constructing hyperedges under HyperAttack. 
Especially, we focus on the modeling method based on KNN and we set experiments to investigate different robustness of parameter $\bm{K}$.

\subsection{Impact on perturbation}

\begin{figure*}[ht]\label{figure}
    \centering
    \subfigure[Hypergraph\_KNN (Cora)]{\includegraphics[width=0.32\linewidth]{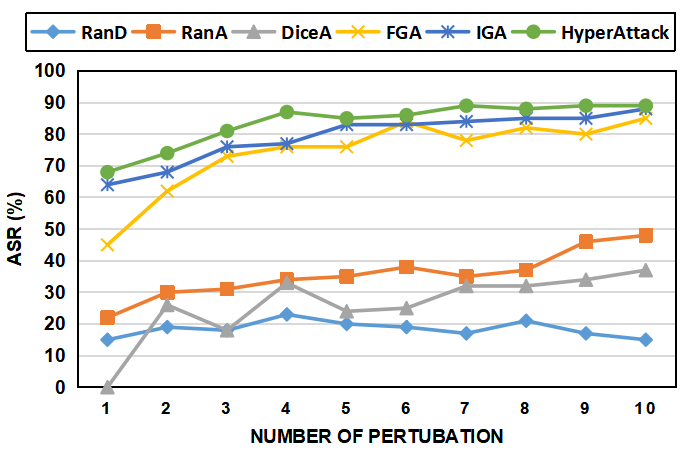}}
    \subfigure[Hypergraph\_$\epsilon$ (Cora)]{\includegraphics[width=0.32\linewidth]{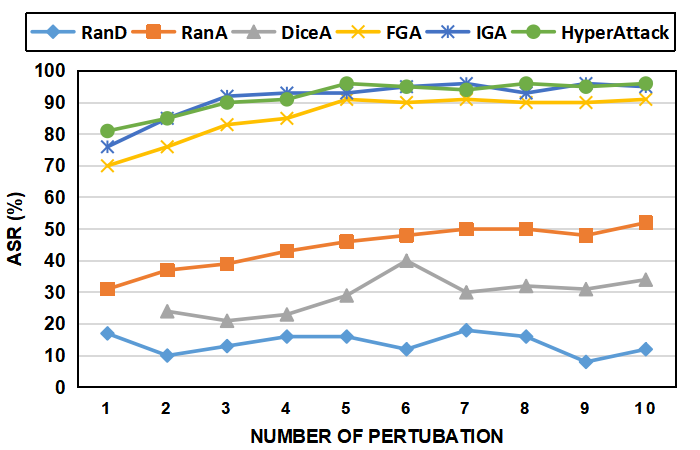}}
    \subfigure[Hypergraph\_L1 (Cora)]{\includegraphics[width=0.32\linewidth]{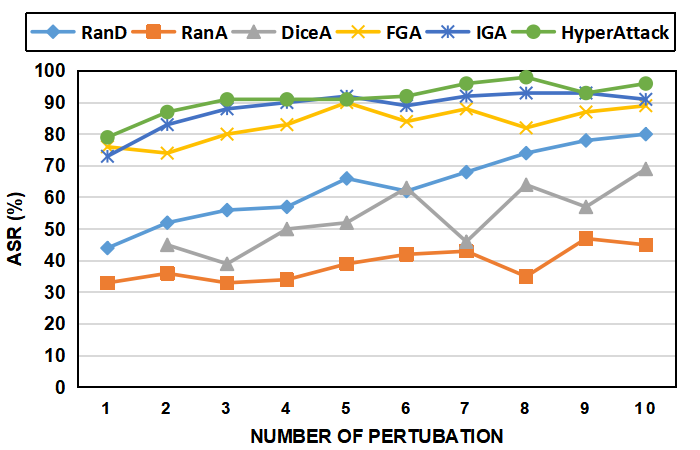}}
    \subfigure[Hypergraph\_KNN (Pubmed)]{\includegraphics[width=0.32\linewidth]{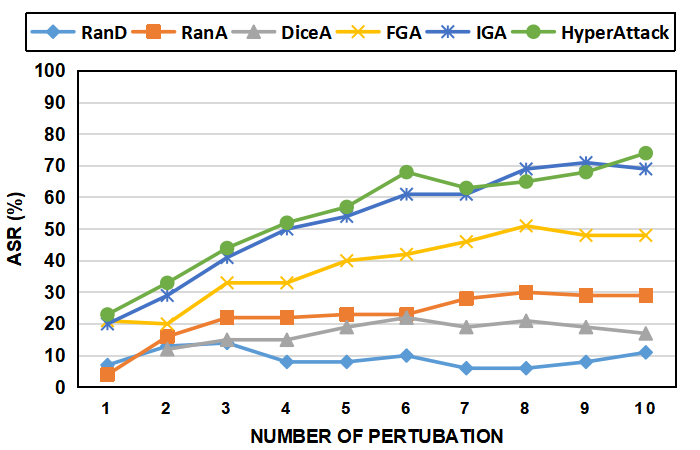}}
    \subfigure[Hypergraph\_$\epsilon$ (Pubmed)]{\includegraphics[width=0.32\linewidth]{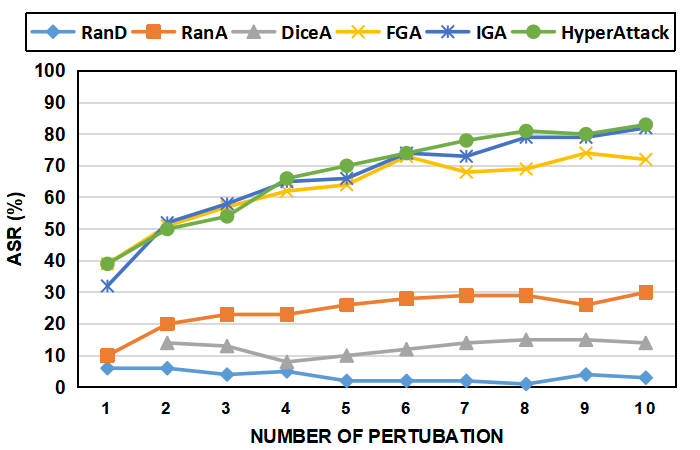}}
    \subfigure[Hypergraph\_L1 (Pubmed)]{\includegraphics[width=0.32\linewidth]{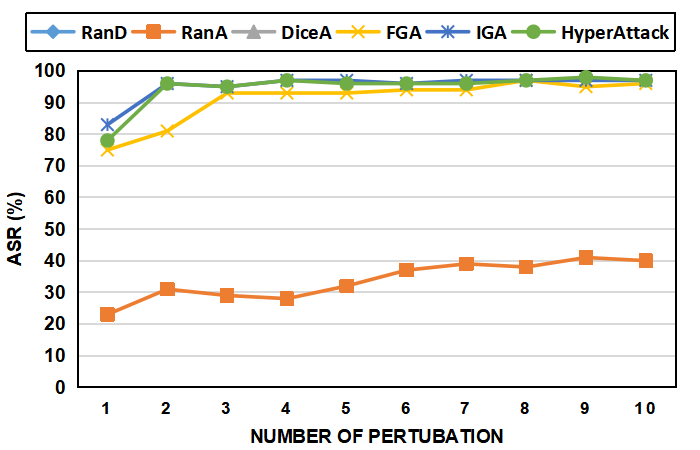}}
    \caption{ASR of different attack methods as functions of perturbation size $\gamma$  on various hypergraph generation methods.}
    \label{fig_pertu_asr}
\end{figure*}

Figure~\ref{fig_pertu_asr} show the results on ASR of different attack methods as functions of perturbation size $\gamma$ on various hypergraph model for the Cora dataset and PubMed dataset.
In the Random-based attack, ASR is not proportional to the number of perturbations due to the high-level randomness in the methods.
For example, as shown in Figure~\ref{fig_pertu_asr} (c), the ASR of DiceA is fluctuating as $\gamma$ increases.
In the Gradient-based attack, ASR shows an overall growing tendency as the number of perturbations rises.
When $\gamma$ increases to a certain value, ASR may reach an upper limit and no longer change significantly, as shown in Figure~\ref{fig_pertu_asr} (f).

\begin{figure}[t]
\centering
\includegraphics[width=0.8\linewidth]{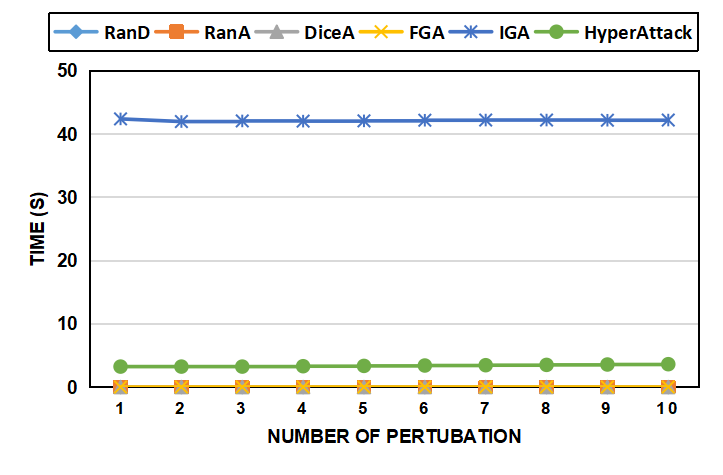}
\caption{Time cost of attack methods with different perturbation size $\gamma$ on Cora dataset.}
\label{fig:Running Time}
\vspace{-0.4cm} 
\end{figure}

Figure~\ref{fig:Running Time} shows the Running Time of the attack with an increasing number of perturbations.
The running time of the attack consists of the time to modify the incidence matrix and the time to calculate the gradient.
The time consumption for modifying the incidence is very small, so the number of perturbations has little effect on the Random-based attack methods.
As the number of perturbations increases, FGA and Hyperattack calculate more gradients, so RT will increase.
For IGA, it needs to calculate the gradient of all hyperedges associated with the target node regardless of the number of perturbations, therefore, the size of perturbations barely affects it.

\subsection{Impact of hypergraph modeling}

We study the impact of different hypergraph generation methods on the attack. 
As shown in Figure~\ref{fig_pertu_asr}, we find that the Hypergraph\_KNN has the lowest ASR and the Hypergraph\_L1 has the highest ASR on both datasets, which we believe is also related to the number of nodes connected by the hyperedges.
The Hypergraph\_KNN has the largest number of nodes connected to each hyperedge on average, and it is the most stable. The number of nodes connected by each hyperedge of Hypergraph\_L1 is the lowest, it is the most likely to be attacked successfully.
But in terms of classification accuracy, the performance of the Hypergraph\_L1 is optimal.
The topic of how to balance hypergraph performance and stability is important.

\begin{figure}
\centering
\includegraphics[width=.8\linewidth]{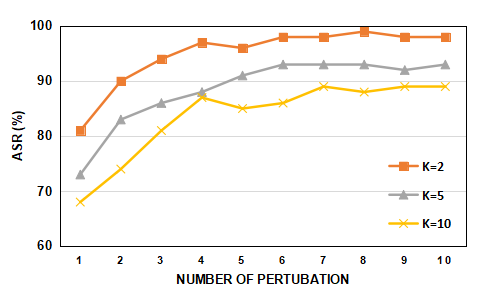}
\caption{ASR of HyperAttack on attacking HGNN and KNN-based  hypergraph construction with different K value.}
\label{fig:discussion}
\vspace{-0.4cm} 
\end{figure}

In particular, we study the effect of hyperedges connected to different numbers of nodes on Hypergraph robustness.
We set experiments and give the results in Figure~\ref{fig:discussion}, which clearly shows the ASR of HyperAttack when the incidence matrix $H$ is constructed based on KNN with different $K$.
We find that the ASR is higher when the $K$ is smaller. 
Therefore, we believe that the stability of the hypergraph is related to the number of nodes connected by the hyperedges,
The hypergraph structure is more stable and robust when the number of nodes connected by the hyperedges is higher.
Meanwhile, when $K=2$, i.e., each hyperedge connects 2 nodes, the constructed hypergraph is a graph structure, indicating that our method is applicable to graph, and has generalization ability.

\section{Related Work}\label{sec7}
In this section, we introduce the related work of hypergraph learning application and the adversarial attack against graph data.

\subsection{Hypergraph learning}
A hypergraph denoted as $\mathcal{G}$, is composed of a set of nodes $\mathcal{V}$, a set of hyperedges $\mathcal{E}$, and a weight matrix $\mathcal{W}$, which represents the connection strengths between the nodes in the hypergraph.

To preserve the high-order associations within the hypergraph and perform downstream tasks effectively, two main steps are involved.
The first step is to construct the hypergraph from the original graph-based data.
Previous studies on hypergraph generation methods have been reviewed in ~\cite{gao2020hypergraph}, which categorize these methods into four types: distance-based ~\cite{HuangLM09,Gao123DObject}, representation-based ~\cite{WangLW15,liu2016elastic,jin2019l2}, attribute-based ~\cite{Huang15,joslyn2019high}, and network-based ~\cite{fang2014topic,zu2016identifying}.
These methods can be further classified into implicit and explicit methods.
Implicit methods, such as distance-based and representation-based methods, do not directly obtain the hyperedges from the original data and require reconstruction through specific measurement and representation algorithms.
Explicit methods, including attribute-based and network-based methods, can directly construct the hyperedges based on the attributes or network connections of the original data, preserving the correlations within the data.

The second step is to design learning methods for the constructed hypergraph. Hypergraph learning can be divided into spectral-analysis methods, neural-network methods, and other methods based on their implementation. 
Spectral-analysis methods are the mainstream in hypergraph learning, utilizing matrix analysis and spectral theory. For instance, \citet{yadati2019hypergcn} employed group expansion to convert the hypergraph into an ordinary graph and used trainable hyperedge perception and hyperedge scoring layers to retain the high-order associativity between nodes and hyperedges. With the advancement of GNN research, some researchers have introduced neural networks into hypergraphs. 
Inspired by graph neural networks, \citet{feng2019hgnn} introduced the Laplace matrix into hypergraphs and proposed the Hypergraph Neural Network (HGNN) framework, which generalizes the star expansion process of hypergraphs to neural networks. 
Other hypergraph learning methods focus on specific applications, such as video images and other fields. For example, \citet{su2017vertex} introduced weighted hypergraphs into the 3D target classification task, allowing for the re-evaluation of node correlations through the weight matrix and obtaining the potential correlations between nodes.

%

\subsection{Adversarial attack on graph data}
In recent years, a large number of researchers have tried to generate adversarial examples to make the graph-based deep learning systems have the ability to resist the input of adversarial test~\cite{YinLSWC23,LiuZCLL22,chen2018fast,Alek2018Toplogy,zugner2018adversarial,Dai18AAttack}.
The vast majority of attack methods realize effective performance by modifying graph structures.
The structure attack manipulates the adjacent matrix of the target node in directions and leads to large classification loss.
Many methods try to apply the traditional software testing technology to graph-based deep learning systems testing via white-box testing and black-box testing.
In white-box settings, attackers are able to acquire internal parameters and modify the graph structure using specific algorithms via these available parameters.
Some gradient-based strategies are proposed to realize effective structure attacks.
Nettack~\cite{zugner2018adversarial} is regarded as the first work to exploit the ideas.
It obtains the approximate optimal attack of a disturbance based on a greedy algorithm by calculating scores for structural attacks and feature attacks.
Some works~\cite{chen2018fast} use a gradient of pairwise nodes based on the target model and select the pair of nodes with a maximum absolute gradient to update the modified adjacency matrix.
Topology attack~\cite{Alek2018Toplogy} uses randomization sampling to select sub-optimal binary perturbations.
Meta-gradient is introduced for the first time~\cite{ZugnerG19Mettack} on graph adjacency matrix to solve a bi-level optimization problem for the poisoning attack.
Mettack greedily adds perturbations via the selected largest meta-gradient.

In black-box settings, the attacker is restricted to propose adversarial samples without any access to the target model.
This kind of attack is the most dangerous attack among the attack methods.
Because some evil attacks can attack the model in the reality with very limited acknowledgment.
Some strategies~\cite{Chang20ARest,Dai18AAttack,WDWT21,Raman021} were proposed to learn the generalized attack policy while only requiring prediction labels from the target classifier.
The attackers regard the attack as a Markov decision process. 
The current decision is only based on the current state and has nothing to do with the previous decision. 
The reward function is designed according to the feedback of the victim model after the attack, and then it is solved by reinforcement learning.

\section{Conclusion}\label{sec8}
In this paper, we propose HyperAttack, the first white-box structure attack framework against hypergraph neural networks.
HyperAttack can modify the incidence matrix of the hypergraph by adding or deleting the target node on specific hyperedges. 
To measure the priority of each hyperedge, we use both gradient and the integrated gradient as indicators.
We conduct a large number of experiments by using five different baseline methods.
In two widely-used datasets Cora and PubMed, HyperAttack can greatly shorten the time of structure attack when the attack effect still achieves state-of-the-art performance at the same time.
We also implement HyperAttack on the hypergraph of three different modeling methods to evaluate the different robustness.
We expect to study other attacks with different levels of knowledge(i.e. black-box attack) on HGNNs, and try to influence the classification result by modifying the node feature matrix in the future.
\normalem
\bibliographystyle{ACM-Reference-Format}
\bibliography{ref.bib}
 
\end{document}